\documentclass[conference]{IEEEtran}

\usepackage{xcolor}
\usepackage{graphicx}
\usepackage{subcaption}
\usepackage{amsmath}
\usepackage{amsfonts}
\usepackage{booktabs}
\usepackage[hidelinks]{hyperref}
\usepackage[mode=buildnew]{standalone}
\usepackage{tikz,tikz-3dplot}
\usetikzlibrary{patterns}
\usepackage{pgfplots}
\pgfplotsset{compat=1.16} 
\graphicspath{{./figures/}}
\usepackage{subcaption}
\definecolor{tabBlue}{HTML}{1f77b4}
\definecolor{tabOrange}{HTML}{ff7f0e}
\definecolor{tabGreen}{HTML}{2ca02c}
\definecolor{tabRed}{HTML}{d62728}
\definecolor{tabPurple}{HTML}{9467bd}
\definecolor{tabBrown}{HTML}{8c564b}
\definecolor{tabPink}{HTML}{e377c2}
\definecolor{tabGray}{HTML}{7f7f7f}
\definecolor{tabOlive}{HTML}{bcbd22}
\definecolor{tabCyan}{HTML}{17becf}

\usepackage{fancyhdr}

\fancypagestyle{specialfooter}{%
  \fancyhf{}
  
  \fancyfoot[L]{Preprint accepted at the International Joint Conference on Neural Networks (IJCNN) 2022}
}

\begin{document}
\title{Canonical convolutional neural networks}

\author{\IEEEauthorblockN{Lokesh Veeramacheneni}
		\IEEEauthorblockA{\textit{ Department of Computer Science} \\
		\textit{Hochschule Bonn-Rhein-Sieg} \\
			}
		\IEEEauthorblockA{\textit{Fraunhofer Center for Machine Learning and SCAI} \\
		lokesh.veeramacheneni@smail.inf.h-brs.de
		}
	    \and
		\IEEEauthorblockN{Moritz Wolter}
		\IEEEauthorblockA{\textit{ High Performance Computing and Analytics Lab} \\
		\textit{University of Bonn} \\
			}
		\IEEEauthorblockA{\textit{Fraunhofer Center for Machine Learning and SCAI} \\
		moritz.wolter@uni-bonn.de
		} \\
		\and
		\IEEEauthorblockN{ \hspace{2.6cm} Reinhard Klein}
		\IEEEauthorblockA{\hspace{2.6cm} \textit{Department of Computer Science} \\
		\hspace{2.6cm} \textit{University of Bonn} \\
		\hspace{2.6cm} rk@cs.uni-bonn.de
			}
		\and
		\IEEEauthorblockN{\hspace{2.65cm} Jochen Garcke}
		\IEEEauthorblockA{\hspace{2.65cm} \textit{Fraunhofer Center for Machine Learning and SCAI} \\
				}
		\IEEEauthorblockA{\hspace{2.65cm} \textit{Institute for Numerical Simulation} \\
		\hspace{2.65cm} \textit{University of Bonn} \\
		\hspace{2.65cm} garcke@ins.uni-bonn.de, jochen.garcke@scai.fhg.de
			}
		
		}


\date{\today}

\maketitle

\begin{abstract}
We introduce canonical weight normalization for convolutional neural networks. 
Inspired by the canonical tensor decomposition, we express the weight tensors in so-called canonical networks as scaled sums of outer vector products. In particular, we train network weights in the decomposed form, where scale weights are optimized separately for each mode.
Additionally, similarly to weight normalization, we include a global scaling parameter. We study the initialization of the canonical form by running the power method and by drawing randomly from Gaussian or uniform distributions.
Our results indicate that we can replace the power method with cheaper initializations drawn from standard distributions.
The canonical re-parametrization leads to competitive normalization performance on the MNIST, CIFAR10, and SVHN data sets.
Moreover, the formulation simplifies network compression.
Once training has converged, the canonical form allows convenient model-compression by truncating the parameter sums.
\end{abstract}

\thispagestyle{specialfooter}
\section{Introduction}\label{sec:intro}
While deep neural networks have increased in size and complexity, the tensor structure of convolutional kernels and weight matrices has not changed as rapidly. We believe that much of the potential that tensor representations, such as so-called canonical decompositions, can offer remains to be discovered.

This paper proposes to express, train, \textit{normalize, and compress} network weight tensors in a canonical form as a sum of weighted normalized outer vector products. Similar to weight normalization \cite{Salimans2016Weight}, the resulting canonical normalization allows us to learn an overall length weight. 
In addition to the overall length, the canonical representation lets the optimizer scale all individual rank terms in the sum separately, where all canonical rank vectors are re-normalized after each optimization step. Consequently, the sum's rank weights remain meaningful throughout training. Orientation and magnitude are decoupled.
In addition to normalization, we observe that the canonical representation is helpful for network compression.
Having learned the scales for each rank separately, we can truncate the sum of outer products according to the weight of each rank component to allow easy compression.

The methods closest to ours are the low-rank form forms proposed in \cite{Lebedev2015SpeedingupCN} and \cite{Tai2016Convolutional}. Like \cite{Tai2016Convolutional}, we overcome the instability problem observed by \cite{Lebedev2015SpeedingupCN}. Instead of working with batch-normalization \cite{Tai2016Convolutional}, we propose CP-normalization.  We jointly consider convolutional and fully connected layers, propose a new way to initialize the canonical form, and explore compression.
CP-normalization is a form of weight normalization  \cite{Salimans2016Weight} for convolutional neural networks. 

We implement the proposed methods and all experiments using Tensorly~\cite{Kossaifi2019TensorLy} and Pytorch~\cite{paszke2017automatic}. Our contributions can be summarized as follows:
\begin{itemize}
\item Canonical convolutional neural networks re-express network weights as sums of outer vector products. The formulation improves convergence by scaling overall and per rank lengths separately.
\item We compare our formulation to weight normalization \cite{Salimans2016Weight} and the low rank form of \cite{Tai2016Convolutional}, on the MNIST \cite{lecun1998gradient}, CIFAR-10 \cite{krizhevsky2009learning}, and SVHN \cite{Netzer2011ReadingDI} data sets. We find CP-normalization performs competitively.
\item Having optimized weights for every rank summand, we can sort the summands according to the absolute value of their weight and truncate them according to importance. Consequently, the CP-formulation allows straightforward weight compression by truncation.
\item We study the initialization of the canonical form and replace the standard power method approach with direct initialization for an AlexNet-like architecture on CIFAR-10.
\end{itemize} 
For reproducibility and future work, the source code is available at \url{https://github.com/Fraunhofer-SCAI/canonical-cnn}.

\section{Related Work} \label{sec:related_work}

\subsection{Normalization and Regularization}
Normalization and regularization methods broadly fall into three categories.
Noise-based methods encourage generalization by corrupting network features or input data. The noise randomly hides certain features, thereby denying overfitting by forcing the network to rely on multiple features to evade the noise. Dropout \cite{srivastava2014dropout}, adaptive dropout \cite{ba2013adaptive}, stochastic pooling \cite{Zeiler2013Stochastic}, input noise \cite{bishop2006pattern,Graves2012Supervised} as well as weight or synaptic noise \cite{Graves2012Supervised} fall into this category. Dropout randomly removes neurons during training, using a fixed removal probability. Adaptive dropout optimizes the removal probability per neuron. Stochastic pooling sub-samples by choosing activations randomly. Input noise randomly corrupts training inputs. Weight noise is added to the parameters to move the model away from local minima and reduce the amount to which subsequent layers can rely on individual features.

Methods in the second group change the cost function to encourage generalization. Weight decay adds an $L_2$ loss term to the cost function \cite{bishop2006pattern,goodfellow2016deep}, to prevent excessive parameter growth. Placing a cost on the $L_2$ parameter norm prevents weight growth and limits network complexity by pushing parameters to zero.  

Finally, structural methods modify the network structure to achieve a regularizing effect. Batch normalization \cite{ioffe2015batch} for example, adjusts the mean and variance of intermediate representations to be approximately standard normal. Weight normalization \cite{Salimans2016Weight} resets the length of weight tensors and introduces an additional length parameter per tensor. It measures tensor length by computing the length of a corresponding flat vector. The normalized weight vector is divided by the vector length after every weight update~\cite{Salimans2016Weight}. 

Weight decay penalizes the $L_2$ term. Normalization does not. It merely seeks to improve the conditioning of the underlying optimization problem.

Our approach also falls into this category. Even though the Euclidean length is a valid tensor norm \cite{Kolda2009Tensor}, we argue that preserving and measuring individual rank length is beneficial because it shares the normalizing properties of weight normalization while additionally simplifying network compression.

\subsection{Network Compression}
Two effective ways to compress artificial neural networks are quantization \cite{KrishnamoorthiQuantization2018} and pruning.
Quantization techniques save memory space and computation time by storing the network tensors at less than floating-point precision.
Pruning removes parameters that contribute little to the overall performance.
Pruned weights are, for example, removed based on the individual magnitude of weights \cite{Han2016DeepCC} or by removing entire rows based on the row-norm \cite{Lebedev2016FastCU} for improved efficiency.

An alternative to pruning is to enforce sparse or structured matrices a-priori. By shifting and reusing the same row, implementing circulant matrix structures saves weights \cite{araujo2018training}. Alternatively, the frequency domain can help us impose sparse diagonal patterns onto the network weight matrices. The fast-food approach proposed in \cite{Yang2015Deep} works with a Welsh-Hadamard transform. In addition to the fixed Welsh-Hadamard transform, adaptive wavelet transforms \cite{Wolter2020Neural} are known to work.

\subsection{Tensor Decompositions}  
Tensor decompositions are well-established in science, and engineering \cite{de2000multilinear,Kolda2009Tensor}. The machine learning community has previously studied CANDECOMP/PARAFAC (CP) decompositions, see, e.g., \cite{Kuzmin2019,Ma2020SpatioTemporalTS}. A common approach is to compress pre-trained convolutional neural networks \cite{astrid2017cp,Jaderberg2014Speeding,Kuzmin2019,Lebedev2015SpeedingupCN,Phan2020Stable}. In particular, after converting the converged weights to the canonical format, \cite{Lebedev2015SpeedingupCN} uses fine-tuning after application of the CP-decomposition, with tiny learning rates. 
\cite{Beylkin.Garcke.Mohlenkamp:2009} investigated sums of separable functions as a functional analog to the canonical decomposition for regression and classification~\cite{Garcke:2010}. \cite{Phan2020Stable} adjusts the computation of the CP-decomposition to yield a representation that is stable during the ensuing fine-tuning.

An alternative to the canonical or parallel factors format is the tensor train representation~\cite{Kolda2009Tensor}. The tensor train format has been used to train compressed versions of fully connected layers in CNN \cite{Novikov2015TensorizingNN,WU2020309}, RNN \cite{tjandra2017compressing}, and GANs~\cite{obukhov2021spectral}. 

Training low-rank CNN from scratch was previously explored in \cite{Tai2016Convolutional}. The proposed approach introduces additional layers. The extra layers lead to deeper networks, which are harder to train. Batch normalization is applied to deal with arising instabilities. 

We argue that introducing additional layers is not required. Our scaled CP-coefficients stabilize training and allow joint normalization and compression similar to weight normalization.

To bolster our argument, we study the link between normalization and compression. We, therefore, apply a canonical formulation already during training.
To stabilize the formulation, we regularly re-normalize our vectors. We explore initialization by computing the CP-decomposition and random initializations using various distributions. 
Additionally, we will explore truncating the canonical sum for compression after convergence.

\section{Methods} \label{sec:methods}
In this section, we will discuss the mathematical background and notations.
Afterwards, we briefly revisit the canonical tensor decomposition and introduce our weight re-parametrization. We adopt the notation from \cite{Kolda2009Tensor}, to which we also refer for a further introduction into tensor decompositions. Throughout the text, vectors are denoted as boldface lowercase letters, for example, $\mathbf{x}, \mathbf{y}$. Capital letters $\mathbf{A}, \mathbf{B}$ stand for matrices. Finally, Euler script $\mathcal{X}$ denotes tensors.

\subsection{The Outer Product $\circ $}
In the two dimensional case, the outer product $\circ $ of two vectors $x \in \mathbb{R}^{i}, y \in \mathbb{R}^{j}$,
\begin{align}
\mathbf{x} \circ \mathbf{y} = \mathbf{x}\mathbf{y}^{T} = \mathbf{A},
\end{align}
results in a matrix $\mathbf{A} \in \mathbb{R}^{i,j}$. If we wanted to turn $\mathbf{A}$ into tensor $\mathcal{A}$ we could simply add additional vectors to the chain of outer products. For example, using $\mathbf{z} \in \mathbb{R}^{k}$ we could produce $\mathbf{x} \circ \mathbf{y} \circ \mathbf{z} = \mathcal{A} \in \mathbb{R}^{i,j,k}$. Formally speaking, a $n$-dimensional tensor $\mathcal{X} \in \mathbb{R}^{d_1, \dots, d_n}$ \textit{of rank one} can be rewritten as an outer product of $n$ vectors \cite{Kolda2009Tensor}
\begin{align}
	\label{eq:one_rank_tensor}
\mathcal{X}=\mathbf{x}^{(1)}\circ \mathbf{x}^{(2)} \circ \mathbf{x}^{(3)} ... \circ \mathbf{x}^{(n)},	
\end{align}
where $\mathcal{X}$ and $\mathbf{x}^{(1)}, \mathbf{x}^{(2)}, \mathbf{x}^{(3)}, \ldots, \mathbf{x}^{(n)}$ are tensor and vectors respectively.
The bracket powers denote series elements. The vectors have the size of the dimension at their position in the sequence from Eq. (\ref{eq:one_rank_tensor}). 

For an individual element in $\mathcal{X}$ at position $i_1, i_2, 
\ldots, i_n$ in the tensor this means that \cite{Kolda2009Tensor},
\begin{align}
	x_{i_1, i_2, \dots , i_n} = x^{(1)}_{i_1} x^{(2)}_{i_2} x^{(3)}_{i_3} \dots x^{(n)}_{i_n}.
\end{align}

\subsection{CP-Decomposition}
\begin{figure}
	\centering
	  \tdplotsetmaincoords{70}{115}

  \begin{tikzpicture}[scale=3,tdplot_main_coords]
    \coordinate (O) at (0,0,0);
    \coordinate (Px) at (1,0,0);
    \coordinate (Py) at (0,1,0);
    \coordinate (Pz) at (0,0,1);
    \coordinate (Pxy) at (1,1,0);
    \coordinate (Pxz) at (1,0,1);
    \coordinate (Pyz) at (0,1,1);
    \coordinate (P) at (1,1,1);
    

    \draw[fill=gray!50,fill opacity=0.1] (O) -- (Py) -- (Pyz) -- (Pz) -- cycle;
    \draw[fill=gray!50,fill opacity=0.1] (O) -- (Px) -- (Pxy) -- (Py) -- cycle;
    \draw[fill=gray!50,fill opacity=0.1] (O) -- (Px) -- (Pxz) -- (Pz) -- cycle;
    \draw[pattern color=tabBlue, pattern=north east lines] (Pz) -- (Pyz) -- (P) -- (Pxz) -- cycle;
    \draw[pattern color=tabRed, pattern=horizontal lines] (Px) -- (Pxy) -- (P) -- (Pxz) -- cycle;
    \draw[pattern color=tabGreen, pattern=vertical lines] (Py) -- (Pxy) -- (P) -- (Pyz) -- cycle;

    \node[text width=1cm] at (1, 1.8, 1) (equal1){$=$};
    \node[text width=.4cm] at (1, 2.05, 1.) (lambda1){\Large $\lambda_1$};

    \draw[pattern color=tabBlue, pattern=north east lines]  (1, 2, 1.1) -- (0, 2, 1.1) -- (0, 2.2, 1.1) -- (1, 2.2, 1.1) -- cycle;
    \draw[pattern color=tabRed, pattern=horizontal lines] (1, 2, 0.9) -- (1, 2, 0) -- (1, 2.2, 0) -- (1, 2.2, 0.9) -- cycle;
    \draw[pattern color=tabGreen, pattern=vertical lines] (1, 2.2, 1.1) -- (1, 3, 1.1) -- (1, 3, .9) -- (1, 2.2, .9) -- cycle;
    \node[text width=.4cm] at (1., 2.3, 0.1) (a1){\Large $\mathbf{a}_1$};
    \node[text width=.4cm] at (1., 2.9, 0.8) (b1){\Large $\mathbf{b}_1$};
    \node[text width=.4cm] at (0., 2.2, 0.98) (c1){\Large $\mathbf{c}_1$};

    \node[text width=.3cm] at (1, 3.1, 1.) (plus1){$+$};
    \node[text width=.4cm] at (1, 3.25, 1.) (lambda2){\Large $\lambda_2$};
    
    \draw[pattern color=tabBlue, pattern=north east lines]  (1, 3.2, 1.1) -- (0, 3.2, 1.1) -- (0, 3.4, 1.1) -- (1, 3.4, 1.1) -- cycle;
    \draw[pattern color=tabRed, pattern=horizontal lines] (1, 3.2, 0.9) -- (1, 3.2, 0) -- (1, 3.4, 0) -- (1, 3.4, 0.9) -- cycle;
    \draw[pattern color=tabGreen, pattern=vertical lines] (1, 3.4, 1.1) -- (1, 4.2, 1.1) -- (1, 4.2, .9) -- (1, 3.4, .9) -- cycle;
    \node[text width=.4cm] at (1., 3.5, 0.1) (a2){\Large $\mathbf{a}_2$};
    \node[text width=.4cm] at (1., 4.1, 0.8) (b2){\Large $\mathbf{b}_2$};
    \node[text width=.4cm] at (0., 3.4, 0.98) (c2){\Large $\mathbf{c}_2$};

    \node[text width=.3cm] at (1, 4.35, 1.) (plus2){$+$};
    \node[text width=.4cm] at (1, 4.55, 1.) (dots){\Large $\dots$};

    \node[text width=.4cm] at (1, 4.8, 1.) (plus3){$+$};
    \node[text width=.4cm] at (1, 5.05, 1.) (lambda2){\Large $\lambda_n$};
    \draw[pattern color=tabBlue, pattern=north east lines]  (1, 5, 1.1) -- (0, 5, 1.1) -- (0, 5.2, 1.1) -- (1, 5.2, 1.1) -- cycle;
    \draw[pattern color=tabRed, pattern=horizontal lines] (1, 5, 0.9) -- (1, 5, 0) -- (1, 5.2, 0) -- (1, 5.2, 0.9) -- cycle;
    \draw[pattern color=tabGreen, pattern=vertical lines] (1, 5.2, 1.1) -- (1, 6, 1.1) -- (1, 6, .9) -- (1, 5.2, .9) -- cycle;    
    \node[text width=.4cm] at (1., 5.3, 0.1) (an){\Large $\mathbf{a}_n$};
    \node[text width=.4cm] at (1., 5.9, 0.8) (bn){\Large $\mathbf{b}_n$};
    \node[text width=.4cm] at (0., 5.2, 0.98) (cn){\Large $\mathbf{c}_n$};

\end{tikzpicture}
	\caption{Visualization of a canonical decomposition of a tensor in $\mathbb{R}^3$. For a third order tensor of size $\mathbb{R}^{i,j,k}$ we expect three vectors of shape $\mathbf{a} \in \mathbb{R}^i, \mathbf{b} \in \mathbb{R}^j, \mathbf{c} \in \mathbb{R}^k $ in the outer products of each summand. Vectors are expressed by colored rectangles. Orthogonally arranged rectangles symbolize outer products. Assuming unit length vectors, we include $\lambda_r$ in each element of the sum. The summation runs until $n$ or the total CP-tensor rank is reached.}
	\label{fig:CP}
\end{figure}
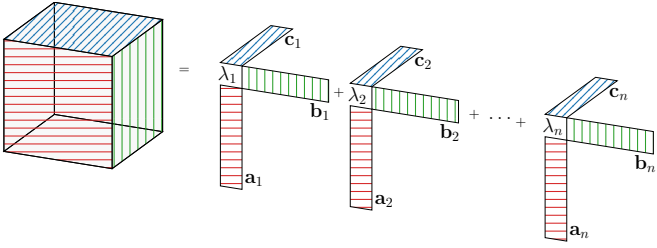
The CP-decomposition expresses a tensor $\mathbf{\mathcal{X}}$ as a sum of $R$ rank one tensors \cite{Kolda2009Tensor}. A three dimensional tensor requires three vectors in each rank product. Consequently
\begin{align}
     \mathbf{\mathcal{X}} \approx [\mathbf{A}, \mathbf{B}, \mathbf{C}]
               = \sum_{r=1}^{R} \mathbf{a}_r \circ \mathbf{b}_r \circ \mathbf{c}_r
\end{align}
approximates the tensor.
Adding an additional scaling weight $\lambda$~\cite{Kolda2009Tensor} allows normalizing the vectors to have unit length in the two-norm, we obtain
\begin{align} \label{eq:cp_form}
    \mathbf{\mathcal{X}} \approx [\mathbf{A}, \mathbf{B}, \mathbf{C}]
    = \sum_{r=1}^{R} \lambda_r \mathbf{a}_r \circ \mathbf{b}_r \circ \mathbf{c}_r,
\end{align}
where $\mathbf{\mathcal{X}}$ is the input tensor and $\mathbf{a}_r,\mathbf{b}_r,\mathbf{c}_r$ are the used vectors representing it, see Figure~\ref{fig:CP}.
The matrices $\mathbf{A}, \mathbf{B}, \mathbf{C}$ contain the vectors $\mathbf{a}_r, \mathbf{b}_r, \mathbf{c}_r$ in their columns. 

Multiple algorithms exist to obtain the CP-form of a tensor: 
Alternating Least Squares (ALS) \cite{Kolda2009Tensor}, the
tensor power method \cite{Wang2015FastAG} and Non-linear Least Squares (NLS) \cite{cp_nls}.
We compare the ALS and the power method to random initialization for our networks.
The ALS approach iteratively and alternatingly updates
$\mathbf{A}, \mathbf{B}, \mathbf{C}$, i.e. separately each mode matrix of the tensor. 
The power method starts from a random initialization and relies on repeated multiplications to find the CP-decomposition.
The procedure is similar to the matrix case. We refer the interested reader to \cite{Kolda2009Tensor}.
Once initialized, we optimize our networks in the CP-form.

\subsection{Canonical Weight Normalization (CPNorm)}\label{sec:canonical_norm}
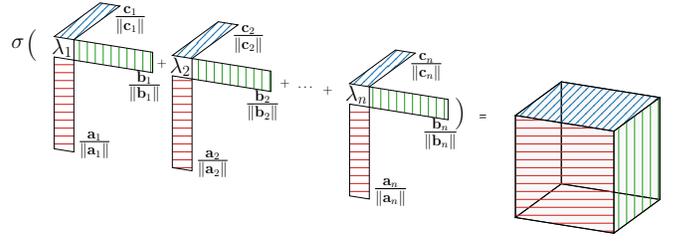
\begin{figure}
	\centering
	  \tdplotsetmaincoords{70}{115}

  \begin{tikzpicture}[scale=3,tdplot_main_coords]
    \coordinate (O) at (1.75,7.5,1.0-.25);
    \coordinate (Px) at (2.75,7.5,1.0-.25);
    \coordinate (Py) at (1.75,8.5,1.0-.25);
    \coordinate (Pz) at (1.75,7.5,2.0-.25);
    \coordinate (Pxy) at (2.75,8.5,1.0-.25);
    \coordinate (Pxz) at (2.75,7.5,2.0-.25);
    \coordinate (Pyz) at (1.75,8.5,2.0-.25);
    \coordinate (P) at (2.75,8.5,2.0-.25);
    

    \draw[fill=gray!50,fill opacity=0.1] (O) -- (Py) -- (Pyz) -- (Pz) -- cycle;
    \draw[fill=gray!50,fill opacity=0.1] (O) -- (Px) -- (Pxy) -- (Py) -- cycle;
    \draw[fill=gray!50,fill opacity=0.1] (O) -- (Px) -- (Pxz) -- (Pz) -- cycle;
    \draw[pattern color=tabBlue, pattern=north east lines] (Pz) -- (Pyz) -- (P) -- (Pxz) -- cycle;
    \draw[pattern color=tabRed, pattern=horizontal lines] (Px) -- (Pxy) -- (P) -- (Pxz) -- cycle;
    \draw[pattern color=tabGreen, pattern=vertical lines] (Py) -- (Pxy) -- (P) -- (Pyz) -- cycle;
    \node[text width=1cm] at (1, 1.75, 1) (sigma){\LARGE $\sigma$};
    \node[text width=1cm] at (1, 1.9, 1) (bracketopen){\Huge$($};
    \node[text width=.4cm] at (1, 2.05, 1.) (lambda1){\LARGE $\lambda_1$};

    \draw[pattern color=tabBlue, pattern=north east lines]  (1, 2, 1.1) -- (0, 2, 1.1) -- (0, 2.2, 1.1) -- (1, 2.2, 1.1) -- cycle;
    \draw[pattern color=tabRed, pattern=horizontal lines] (1, 2, 0.9) -- (1, 2, 0) -- (1, 2.2, 0) -- (1, 2.2, 0.9) -- cycle;
    \draw[pattern color=tabGreen, pattern=vertical lines] (1, 2.2, 1.1) -- (1, 3, 1.1) -- (1, 3, .9) -- (1, 2.2, .9) -- cycle;
    \node[text width=.85cm] at (1., 2.4, 0.1) (a1){\LARGE $\frac{\mathbf{a}_1}{\| \mathbf{a}_1 \|}$};
    \node[text width=.85cm] at (1., 2.9, 0.75) (b1){\LARGE $\frac{\mathbf{b}_1}{\| \mathbf{b}_1\|}$};
    \node[text width=.85cm] at (0., 2.3, 0.98) (c1){\LARGE $\frac{\mathbf{c}_1}{\| \mathbf{c}_1\|}$};

    \node[text width=.3cm] at (1, 3.1, 1.) (plus1){$+$};
    \node[text width=.4cm] at (1, 3.25, 1.) (lambda2){\LARGE $\lambda_2$};
    
    \draw[pattern color=tabBlue, pattern=north east lines]  (1, 3.2, 1.1) -- (0, 3.2, 1.1) -- (0, 3.4, 1.1) -- (1, 3.4, 1.1) -- cycle;
    \draw[pattern color=tabRed, pattern=horizontal lines] (1, 3.2, 0.9) -- (1, 3.2, 0) -- (1, 3.4, 0) -- (1, 3.4, 0.9) -- cycle;
    \draw[pattern color=tabGreen, pattern=vertical lines] (1, 3.4, 1.1) -- (1, 4.2, 1.1) -- (1, 4.2, .9) -- (1, 3.4, .9) -- cycle;
    \node[text width=.85cm] at (1., 3.6, 0.1) (a2){\LARGE $\frac{\mathbf{a}_2}{\|\mathbf{a}_2\|}$};
    \node[text width=.85cm] at (1., 4.1, 0.75) (b2){\LARGE $\frac{\mathbf{b}_2}{\|\mathbf{b}_2\|}$};
    \node[text width=.85cm] at (0., 3.5, 0.98) (c2){\LARGE $\frac{\mathbf{c}_2}{\|\mathbf{c}_2\|}$};

    \node[text width=.3cm] at (1, 4.35, 1.) (plus2){$+$};
    \node[text width=.4cm] at (1, 4.55, 1.) (dots){$\dots$};

    \node[text width=.4cm] at (1, 4.8, 1.) (plus3){$+$};
    \node[text width=.5cm] at (1, 5.05, 1.) (lambda2){\LARGE $\lambda_n$};
    \draw[pattern color=tabBlue, pattern=north east lines]  (1, 5, 1.1) -- (0, 5, 1.1) -- (0, 5.2, 1.1) -- (1, 5.2, 1.1) -- cycle;
    \draw[pattern color=tabRed, pattern=horizontal lines] (1, 5, 0.9) -- (1, 5, 0) -- (1, 5.2, 0) -- (1, 5.2, 0.9) -- cycle;
    \draw[pattern color=tabGreen, pattern=vertical lines] (1, 5.2, 1.1) -- (1, 6, 1.1) -- (1, 6, .9) -- (1, 5.2, .9) -- cycle;    
    \node[text width=.85cm] at (1., 5.4, 0.1) (an){\LARGE $\frac{\mathbf{a}_n}{\|\mathbf{a}_n\|}$};
    \node[text width=.85cm] at (1., 5.9, 0.75) (bn){\LARGE $\frac{\mathbf{b}_n}{\|\mathbf{b}_n \|}$};
    \node[text width=.85cm] at (0., 5.3, 0.98) (cn){\LARGE $\frac{\mathbf{c}_n}{\|\mathbf{c}_n \|}$};

    \node[text width=1cm] at (1, 6.25, 1) (bracketclosed){\Huge$)$};
    \node[text width=1cm] at (1, 6.5, 1.0) (bracketclosed){=};

\end{tikzpicture}
	\caption{Visualization of the proposed canonical weight representation. The tensor weights are expressed as sums of scaled outer products, here illustrated as orthogonal stripes. In addition to the rank scales $\lambda$ we use a global length weight $\sigma$. After every update step, the weight vectors are re-normalized.
	The rank scales $\lambda_r$ and the global scale $\sigma$ allow learning. After convergence, the rank scales let us assess the relative importance of each rank. Compressing the network means discarding the least important terms.}
	\label{fig:cp_form}
\end{figure}
Instead of working with a single flat weight vector and a global length parameter, we aim to conserve the tensor structure. 
Figure~\ref{fig:cp_form} illustrates our alternative approach.
In the figure, outer products appear as orthogonally arranged squares.
This paper, therefore, explores a CP-weight formulation. Taking a cue from weight normalization, we introduce the parameter $\sigma$.
For a $\mathbb{R}^3$ tensor we choose to represent our weights as
\begin{align}
	\mathcal{W} = \sigma \left( 
	\sum_{r=1}^{R} \lambda_r 
	\frac{\mathbf{a}_r}{\|\mathbf{a}_r\|_2}
	\circ \frac{\mathbf{b}_r}{\|\mathbf{b}_r\|_2}
	\circ \frac{\mathbf{c}_r}{\|\mathbf{c}_r\|_2} \;
	\right).
\end{align}
The rank scales $\lambda_r$, the parameter vectors $\mathbf{a}_r, \mathbf{b}_r, \mathbf{c}_r$, as well as the global length $\sigma$, are all optimized. 
In tensor numerics for the CP-decomposition and related approaches, normalized vectors often appear in the outer vector product for numerical stability and convenience~\cite{Beylkin.Garcke.Mohlenkamp:2009,Garcke:2010,Kolda2009Tensor}. Regular renormalization should consequently improve stability.
For an $\mathbb{R}^{3}$ tensor, we divide $\mathbf{a}_r$, $\mathbf{b}_r$, $\mathbf{c}_r$ by their norm after each weight update. In other words: All weight vectors are normalized after each update. Renormalization preserves their unit length. 

Enforcing unit weight vectors ensures that the rank weight $\lambda_r$ scales the rank. We can now optimize the global and the per rank scales separately. 
Furthermore, by sorting the scales, we can truncate the sum and preserve the essential terms.  

We found our weight formulation to be differentiable in PyTorch. The upcoming section will verify its stability and convergence properties empirically.

\section{Experiments}\label{sec:exp}
\begin{figure}
\centering
\begin{subfigure}[c]{0.3\linewidth}
\begin{tikzpicture}

\begin{axis}[
hide x axis,
hide y axis,
tick align=outside,
tick pos=left,
x grid style={white!69.0196078431373!black},
xmin=-0.5, xmax=159.5,
xtick style={color=black},
y dir=reverse,
y grid style={white!69.0196078431373!black},
ymin=-0.5, ymax=159.5,
ytick style={color=black}
]
\addplot graphics [includegraphics cmd=\pgfimage,xmin=-0.5, xmax=159.5, ymin=159.5, ymax=-0.5] {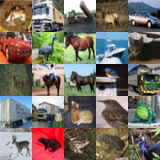};
\end{axis}

\end{tikzpicture}
	\subcaption{CIFAR10}\label{fig:data_CIFAR10}
\end{subfigure}
\begin{subfigure}[c]{0.3\linewidth}
\begin{tikzpicture}

\begin{axis}[
hide x axis,
hide y axis,
tick align=outside,
tick pos=left,
x grid style={white!69.0196078431373!black},
xmin=-0.5, xmax=159.5,
xtick style={color=black},
y dir=reverse,
y grid style={white!69.0196078431373!black},
ymin=-0.5, ymax=159.5,
ytick style={color=black}
]
\addplot graphics [includegraphics cmd=\pgfimage,xmin=-0.5, xmax=159.5, ymin=159.5, ymax=-0.5] {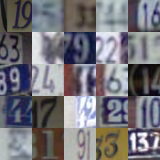};
\end{axis}

\end{tikzpicture}
	\subcaption{SVHN}\label{fig:data_SVHN}
\end{subfigure}
\begin{subfigure}[c]{0.3\linewidth}
\begin{tikzpicture}

\begin{axis}[
hide x axis,
hide y axis,
tick align=outside,
tick pos=left,
x grid style={white!69.0196078431373!black},
xmin=-0.5, xmax=139.5,
xtick style={color=black},
y dir=reverse,
y grid style={white!69.0196078431373!black},
ymin=-0.5, ymax=139.5,
ytick style={color=black}
]
\addplot graphics [includegraphics cmd=\pgfimage,xmin=-0.5, xmax=139.5, ymin=139.5, ymax=-0.5] {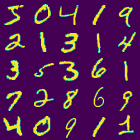};
\end{axis}

\end{tikzpicture}
		\subcaption{MNIST}\label{fig:data_MNIST}
\end{subfigure}
\caption{Visualization of the CIFAR-10~\cite{krizhevsky2009learning}, SVHN~\cite{Netzer2011ReadingDI} and MNIST~\cite{lecun1998gradient} data sets we used to train and evaluate our networks.}\label{fig:data}
\end{figure}

This section describes CP-normalization and compression experiments on LeNet and AlexNet-like architectures. The implementation relies on Pytorch \cite{paszke2017automatic} and TensorLy \cite{Kossaifi2019TensorLy}.
We work with the MNIST \cite{lecun1998gradient}, CIFAR-10~\cite{krizhevsky2009learning}, and SVHN~\cite{Netzer2011ReadingDI} datasets. 

The CIFAR10 dataset consists of 60k images, which we split into 50k train and 10k test images. Ten different categories exist within each split. The networks have to identify planes, cars, birds, cats, deer, dogs, frogs, horses, ships, and trucks. Figure~\ref{fig:data_CIFAR10} shows samples randomly drawn from the training set.

Cropped Street View House Numbers (SVHN) sample images appear in Figure~\ref{fig:data_SVHN}. The dataset contains 73K train and 26K test images. Ten digits from 0 to 9 have to be classified correctly in photos of house numbers obtained from Google Street View.

Finally, the MNIST dataset has 60k elements. Fifty thousand are used to train and 10 to test our networks. Similar to SVHN, handwritten digit numbers have to be classified correctly. Figure~\ref{fig:data_MNIST} shows sample digits.

We train all our networks multiple times to account for the effect of local minima in non-convex optimization.
In total, all architectures in this paper are trained eight times. We report mean values $\mu$ and a single standard deviation $\pm \sigma$ every time.

\subsection{Evaluating Canonical Weight Normalization (CPNorm)}
\subsubsection{AlexNet-CIFAR10}\label{sec:AlexNet-CIFAR10}
This section empirically evaluates canonical normalization using an AlexNet-inspired network. Similar to the classic architecture \cite{NIPS2012_c399862d} we work with five convolutional layers (kernel size-3) and three max pool layers (kernel size-2) followed by three fully connected layers. The last three fully connected layers act as a classifier for the network with dropout applied as in \cite{NIPS2012_c399862d}.

Application of CP-normalization to convolutional or linear layers requires at least approximate prior knowledge of the corresponding layer's rank.

Following \cite{larsen2020practical}, we estimate the tensor-rank of a layer by computing CP-decompositions using the alternating least squares method with increasing ranks until
\begin{align}
1 - \frac{||\mathcal{W}-\bar{\mathcal{W}}||_2}{||\mathcal{W}||_2} \approx 1.
\end{align}
$\mathcal{W}$ represents the original tensor, and $\bar{\mathcal{W}}$ denotes the reconstruction from the CP-form.
\begin{table}[t]
\centering
\caption{Ranks of every layer in our AlexNet-like architecture right after initialization. 4D tensors represent convolutional layers.
2D tensors represent linear layers.
We iteratively estimated the ranks using the alternating least squares (ALS) algorithm.}\label{tab:alexnet_rank}
	\begin{tabular}{ l l } \toprule
		layer size  & tensor rank \\ \midrule 
		3x64x3x3    & 36 \\ 
		64x192x3x3  & 571 \\ 
		192x384x3x3 & 1626 \\ 
		384x256x3x3 & 1948 \\ 
		256x256x3x3	& 1644 \\	
		4096x1024	& 1024 \\
		1024x512	& 512 \\
		512x10		& 10 \\ \bottomrule
	\end{tabular}
\end{table}
Table~\ref{tab:alexnet_rank} lists the full rank of every layer in the AlexNet-like architecture along with every tensor shape.
We estimate the ranks of the initialized tensors before training. 
Patterns and redundant structures are likely to appear during training. The resulting tensors will probably have a lower rank compared to the original tensor.
We accept the larger estimated rank based on the random initialization,
and the limited over-parameterization it causes. 
We believe it tends to help with initial convergence~\cite{Frankle2019Lottery}.
We will revisit this question in Section~\ref{sec:compression}. 
Knowing the rank, we can now convert the weight tensors into their canonical form.

Having established the tensor ranks, we evaluate the stability of the new representation. To this end, we train our AlexNet-like network on CIFAR 10 for 150 epochs. We repeat identical experiments with weight normalization \cite{Salimans2016Weight}, CP-normalization (see section~\ref{sec:canonical_norm}), and without normalization using SGD and RMSprop optimizers. We also apply both weight and canonical normalization to all layers.
The Stochastic Gradient Descent (SGD) optimizer ran with a learning rate of 0.01. We compare the SGD optimization result to multiple runs with an RMSprop optimizer and a learning rate of 0.001.

\begin{table}[t]
\centering
\caption{The test accuracies of an AlexNet inspired network on CIFAR10. We tabulate mean values and a single standard deviation. Experiments without normalization, with weight normalization, and with CP-normalization are compared. For the SGD experiments, we choose a learning rate of 0.01. The RMSProp optimizer ran with a learning rate of 0.001. * represents the early stopping. We find that our CP-normalization approach performs competitively.}
\begin{tabular}{c c c c  c} \toprule
	   &		   & \multicolumn{2}{c}{accuracy [\%]} &			  \\ \cmidrule{3-4}
method & optimizer & max   & $\mu \pm \sigma$ &  weights \\ \midrule
none   & SGD       & 88.01 & $87.06\pm0.36$   & $6.98\times 10^6$ \\ 
weight & SGD	   & 88.63 & $87.38\pm0.37$   & $6.98\times 10^6$ \\ 
CP     & SGD	   & \textbf{89.05} & $\mathbf{88.32\pm0.21}$ & $9.25\times 10^6$ \\ \midrule
none*   & RMSProp   & 83.94 & $82.51\pm1.40$ & $6.98\times 10^6$ \\
weight & RMSProp   & 87.98 & $87.17\pm0.35$ & $6.98\times 10^6$ \\
CP	   & RMSProp   &  \textbf{90.38} & $\mathbf{88.70\pm1.09}$ & $9.25\times 10^6$ \\ \bottomrule
\end{tabular}\label{tab:cifar10_alexnet}

\end{table}
We found empirically that the power method provides better initializations than the alternating least squares approach.
In this and all subsequent experiments, the power method will be used to initialize the weights.

Results are tabulated in Table~\ref{tab:cifar10_alexnet}.
For both SGD and RMSProp, our canonical formulation outperforms weight normalization and the un-normalized network. In the CP-case, we additionally observe faster convergence.
Unfortunately, driving the initial approximation error of the CP-form close to zero increases the network size. We will revisit this issue in section~\ref{sec:compression}.

\subsubsection{AlexNet-SVHN}\label{sec:AlexNet-SVHN}
We now repeat similar experiments on the Street View House Numbers (SVHN) data set. Our AlexNet-like architecture remains the same.
The network is trained for 80 epochs by SGD with a learning rate of 0.08.
Similarly, with RMSProp, we optimize for 80 epochs with a relatively small learning rate of 0.0001. Again the ranks from Table 1 are used to apply CP-normalization. 
\begin{table}[t]
\centering
\caption{Test accuracies of the AlexNet-like network performance on the SVHN dataset.
Experiments with weight-normalization, CP-normalization, and without being shown.
We train using SGD with a learning rate of 0.08. For the RMSProp experiments, we work with a step size of 0.0001.}
\begin{tabular}{c c c c c} \toprule

	   & 		   & \multicolumn{2}{c}{accuracy [\%]}  \\ \cmidrule{3-4}
method & optimizer & max & $\mu \pm \sigma$ &  weights \\ \midrule
none   & SGD       & 95.39 & $94.61\pm0.40$  & $6.98\times 10^6$  \\ 
weight & SGD       & 95.43 & $94.30\pm0.70$   & $6.98\times 10^6$ \\ 
CP     & SGD	   & \textbf{95.62} & $\mathbf{94.82\pm0.29}$  &  $9.25\times 10^6$ \\ \midrule
none   & RMSProp   & 94.76 & $94.00\pm0.33$ & $6.98\times 10^6$  \\
weight & RMSProp   & 94.98 & $94.06\pm0.32$ & $6.98\times 10^6$  \\
CP	   & RMSProp   & \textbf{95.03} & $\mathbf{94.52\pm0.30}$ & $9.25\times 10^6$ \\ \bottomrule
\end{tabular}\label{tab:SVHN_alexnet}
\end{table}

Table~\ref{tab:SVHN_alexnet} shows the performance of our AlexNet-like architecture on the SVHN dataset. Once more, we compare weight- and CP-normalization, as well as no normalization. Once more, CP-normalization improved performance with both optimizers.
Without normalization and higher learning rates, RMSprop was occasionally unstable. Both normalization approaches stabilized these runs successfully.

\subsubsection{LeNet-MNIST}
The implemented LeNet-inspired architecture has two convolutional layers (kernel size-3) followed by two fully connected layers with dropout. 
Just like we did for our previous experiments, we again compare the performance with SGD, and RMSProp on MNIST.
We set the learning rate to 0.001 and train over 50 epochs for both optimizers.
We repeat the rank estimation procedure as described in Section \ref{sec:AlexNet-CIFAR10}, Table~\ref{tab:leNet-Rank} shows the ranks we measured.
\begin{table}
	\centering
	\caption{Ranks of every layer in our LeNet-like architecture.
	Convolutional layers have 4D tensors, and 2D tensor represent linear layers.}\label{tab:leNet-Rank}
	\begin{tabular}{ c c } \toprule
		Layer size  & Tensor rank \\ \midrule 
		1x32x3x3    & 11 \\ 
		32x64x3x3   & 270 \\ 
		9216x128    & 128 \\
		128x10		& 10 \\ \bottomrule
	\end{tabular}
\end{table}
\begin{table}[t]
	\centering
	\caption{Test accuracies of the LeNet-like architecture without normalization, with weight normalization, and with CP-normalization. We compare training with a SGD optimizer to training with RMSProp with a learning rate of 0.001.}
	\label{tab:LeNetMnist}
	\begin{tabular}{c c c c c} 						 \toprule
			  &			  & \multicolumn{2}{c}{accuracy [\%]} &		  \\ \cmidrule{2-4}
		method& optimizer & max & $\mu \pm \sigma$  &  \# weights \\ \midrule
		none  &	SGD		  &  97.91 & $97.79 \pm 0.07$ & $1.19\times 10^6$   \\ 
		weight&	SGD		  & 98.00 & $97.91 \pm 0.07$  & $1.20\times 10^6$      \\ 
		CP    &	SGD		  & \textbf{98.77} & \textbf{$98.66 \pm 0.05$} & $1.22\times 10^6$  \\ \midrule
		none  & RMSProp   & 99.33 & $99.10 \pm 0.05$ &$1.19\times 10^6$ \\
		weight& RMSProp   & \textbf{99.40}& \textbf{$99.30 \pm 0.04$} & $1.20\times 10^6$ \\
		CP	  & RMSProp	  &99.35     &  $99.21 \pm 0.05$ & $1.22\times 10^6$ \\ \bottomrule
	\end{tabular}
\end{table}
Table~\ref{tab:LeNetMnist} depicts the results of LeNet over different normalization methods using SGD and RMSprop optimizers. Similar to the AlexNet-case, LeNet with CP-normalization converges to competitive mean accuracies. The parameter gain caused by the full rank CP-form is less pronounced in this case.

\begin{figure}[t]
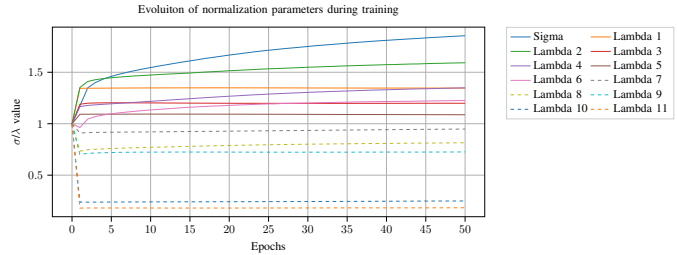

	\centering
		\includestandalone[width=\linewidth]{./figures/plot}
	\caption{Plot of the progression of $\sigma$ and $\lambda$'s during training for the first convolutional layer in the LeNet-like architecture. The first layer has a tensor shape of 1x32x3x3 and a rank of 11.
	Solid lines portray weights with positive growth. We use dashed lines for $\lambda$ values with negative growth.}\label{fig:lambda_plot}
\end{figure}
We show the evolution of the CP-parameters from Equation~\ref{eq:cp_form} in Figure~\ref{fig:lambda_plot}. The $\sigma$ and $\lambda$s are all one initially. We observe that the rate of change for all parameters is initially very high. Towards the end of the training process, all values cease to change. We conclude that gradients are applied, and backpropagation is successful.
Figure~\ref{fig:lambda_plot} also displays significant differences between the various rank weights, a prerequisite for our truncation approach to be meaningful.

\subsection{Random Initialization of the CP-Form}
All networks started in an initialized tensor form thus far. Running the power method converted the original tensor into the CP-form. 
Evaluation of the power method causes additional overhead. This section will explore alternatives to reduce the computational cost. Working with the setup described in Section~\ref{sec:AlexNet-CIFAR10}, we now directly populate the CP-form with random values drawn from standardized distributions.

Table~\ref{tab:alexnet_rank} shows layer sizes of all AlexNet-Layers. Since we store the individual vectors in matrices, the four-dimensional convolutional layers will have four matrices $\mathbf{A}, \mathbf{D}, \mathbf{C}, \mathbf{D}$ per layer. For the fully connected layers, we have two matrices $\mathbf{A}, $ and $\mathbf{B}$. Since the rank of each layer determines the shape, we can construct the CP-shape directly and initialize the matrices using Kaiming normal or uniform distributions. In a first series of experiments we set $\lambda_r = 1$ for all ranks $r$ initially.
\begin{table}[t]
	\centering
	\caption{Test accuracies of the AlexNet-like network performance on the CIFAR10 dataset with various initializations.
	We train using SGD with a learning rate of 0.01 and initially $\lambda_r = 1$ for all ranks. }
	\begin{tabular}{c c c} \toprule
							  &	\multicolumn{2}{c}{accuracy [\%]}  \\ \cmidrule{2-3}
		factor initialization & max     & $\mu \pm \sigma $\\ \midrule
		CP-decomposition      &	89.05	& $88.32 \pm 0.21$ \\ 
		Kaiming normal 		  &	88.95	& $87.76 \pm 0.36$ \\ 
		Kaiming uniform 		  &	88.68	& $87.63 \pm 0.43$ \\ 
		\bottomrule
	\end{tabular}\label{tab:init_alexnet}
\end{table}
Table ~\ref{tab:init_alexnet} compares direct initialization with Kaiming normal or uniformly distributed values to initialization through the CP-decomposition via the power method. We find the performance with stochastic gradient descent competitive. 

An additional series of experiments explores normal instead of constant initialization for the rank-scales  $\lambda$. 
\begin{table}[t]
	\centering
	\caption{Test accuracies of the AlexNet-like network performance on the CIFAR10 dataset with various initializations.
	We train using ADAM with a learning rate of 0.001. A normal distribution $\mathcal{N}(0, 1)$ initialized the rank-scales $\lambda$. }
	\begin{tabular}{c c c} \toprule
							  &	\multicolumn{2}{c}{accuracy [\%]}  \\ \cmidrule{2-3}
		factor initialization & max     & $\mu \pm \sigma $\\ \midrule
		CP-decomposition      &	90.27	& $89.59 \pm 0.21$ \\ 
		Kaiming normal 		  &	90.61	& $89.70 \pm 0.37$ \\ 
		Kaiming uniform 		  &	90.62	& $89.74 \pm 0.35$ \\ 
		\bottomrule
	\end{tabular}\label{tab:init_alexnet_ADAM}
\end{table}
\begin{figure}
	\centering
\begin{tikzpicture}

\definecolor{color0}{rgb}{0.12156862745098,0.466666666666667,0.705882352941177}

\begin{axis}[
title={Initial $\lambda$-distribution},
tick align=outside,
tick pos=left,
x grid style={white!69.0196078431373!black},
xlabel={value},
xmajorgrids,
xmin=-3.73369442224503, xmax=3.73498688936234,
xtick style={color=black},
y grid style={white!69.0196078431373!black},
ylabel={Count},
ymajorgrids,
ymin=0, ymax=183.75,
ytick style={color=black}
]
\draw[draw=none,fill=color0] (axis cs:-3.39420890808105,0) rectangle (axis cs:-3.16788523197174,2);
\draw[draw=none,fill=color0] (axis cs:-3.16788523197174,0) rectangle (axis cs:-2.94156155586243,3);
\draw[draw=none,fill=color0] (axis cs:-2.94156155586243,0) rectangle (axis cs:-2.71523787975311,6);
\draw[draw=none,fill=color0] (axis cs:-2.71523787975311,0) rectangle (axis cs:-2.4889142036438,4);
\draw[draw=none,fill=color0] (axis cs:-2.4889142036438,0) rectangle (axis cs:-2.26259052753448,8);
\draw[draw=none,fill=color0] (axis cs:-2.26259052753448,0) rectangle (axis cs:-2.03626685142517,17);
\draw[draw=none,fill=color0] (axis cs:-2.03626685142517,0) rectangle (axis cs:-1.80994317531586,26);
\draw[draw=none,fill=color0] (axis cs:-1.80994317531586,0) rectangle (axis cs:-1.58361949920654,42);
\draw[draw=none,fill=color0] (axis cs:-1.58361949920654,0) rectangle (axis cs:-1.35729582309723,58);
\draw[draw=none,fill=color0] (axis cs:-1.35729582309723,0) rectangle (axis cs:-1.13097214698792,84);
\draw[draw=none,fill=color0] (axis cs:-1.13097214698792,0) rectangle (axis cs:-0.904648470878601,110);
\draw[draw=none,fill=color0] (axis cs:-0.904648470878601,0) rectangle (axis cs:-0.678324794769287,114);
\draw[draw=none,fill=color0] (axis cs:-0.678324794769287,0) rectangle (axis cs:-0.452001118659973,167);
\draw[draw=none,fill=color0] (axis cs:-0.452001118659973,0) rectangle (axis cs:-0.225677442550659,160);
\draw[draw=none,fill=color0] (axis cs:-0.225677442550659,0) rectangle (axis cs:0.000646233558654785,165);
\draw[draw=none,fill=color0] (axis cs:0.000646233558654785,0) rectangle (axis cs:0.226969909667969,175);
\draw[draw=none,fill=color0] (axis cs:0.226969909667969,0) rectangle (axis cs:0.453293585777283,168);
\draw[draw=none,fill=color0] (axis cs:0.453293585777283,0) rectangle (axis cs:0.679617261886597,148);
\draw[draw=none,fill=color0] (axis cs:0.679617261886597,0) rectangle (axis cs:0.90594093799591,128);
\draw[draw=none,fill=color0] (axis cs:0.90594093799591,0) rectangle (axis cs:1.13226461410522,100);
\draw[draw=none,fill=color0] (axis cs:1.13226461410522,0) rectangle (axis cs:1.35858829021454,84);
\draw[draw=none,fill=color0] (axis cs:1.35858829021454,0) rectangle (axis cs:1.58491196632385,71);
\draw[draw=none,fill=color0] (axis cs:1.58491196632385,0) rectangle (axis cs:1.81123564243317,42);
\draw[draw=none,fill=color0] (axis cs:1.81123564243317,0) rectangle (axis cs:2.03755931854248,32);
\draw[draw=none,fill=color0] (axis cs:2.03755931854248,0) rectangle (axis cs:2.26388299465179,14);
\draw[draw=none,fill=color0] (axis cs:2.26388299465179,0) rectangle (axis cs:2.49020667076111,9);
\draw[draw=none,fill=color0] (axis cs:2.49020667076111,0) rectangle (axis cs:2.71653034687042,5);
\draw[draw=none,fill=color0] (axis cs:2.71653034687042,0) rectangle (axis cs:2.94285402297974,2);
\draw[draw=none,fill=color0] (axis cs:2.94285402297974,0) rectangle (axis cs:3.16917769908905,1);
\draw[draw=none,fill=color0] (axis cs:3.16917769908905,0) rectangle (axis cs:3.39550137519836,3);
\end{axis}

\end{tikzpicture}
	\caption{Histograms depicting initial the initially normal $\mathcal{N}(0, 1)$ distribution of rank-scales ($\lambda$). The histogram shows the fourth layer of our AlexNet-architecture, before a gradient update has been applied.}\label{fig:init_lambda}
\begin{tikzpicture}

\definecolor{color0}{rgb}{0.12156862745098,0.466666666666667,0.705882352941177}

\begin{axis}[
title={Converged $\lambda$-distribution},
tick align=outside,
tick pos=left,
x grid style={white!69.0196078431373!black},
xlabel={value},
xmajorgrids,
xmin=-3.16658330538804, xmax=3.16658330538804,
xtick style={color=black},
y grid style={white!69.0196078431373!black},
ylabel={Count},
ymajorgrids,
ymin=0, ymax=245.7,
ytick style={color=black}
]
\draw[draw=none,fill=color0] (axis cs:-2.87871209580731,0) rectangle (axis cs:-2.68679795608682,1);
\draw[draw=none,fill=color0] (axis cs:-2.68679795608682,0) rectangle (axis cs:-2.49488381636633,1);
\draw[draw=none,fill=color0] (axis cs:-2.49488381636633,0) rectangle (axis cs:-2.30296967664584,2);
\draw[draw=none,fill=color0] (axis cs:-2.30296967664584,0) rectangle (axis cs:-2.11105553692536,7);
\draw[draw=none,fill=color0] (axis cs:-2.11105553692536,0) rectangle (axis cs:-1.91914139720487,11);
\draw[draw=none,fill=color0] (axis cs:-1.91914139720487,0) rectangle (axis cs:-1.72722725748438,20);
\draw[draw=none,fill=color0] (axis cs:-1.72722725748438,0) rectangle (axis cs:-1.5353131177639,26);
\draw[draw=none,fill=color0] (axis cs:-1.5353131177639,0) rectangle (axis cs:-1.34339897804341,33);
\draw[draw=none,fill=color0] (axis cs:-1.34339897804341,0) rectangle (axis cs:-1.15148483832292,100);
\draw[draw=none,fill=color0] (axis cs:-1.15148483832292,0) rectangle (axis cs:-0.959570698602435,111);
\draw[draw=none,fill=color0] (axis cs:-0.959570698602435,0) rectangle (axis cs:-0.767656558881948,126);
\draw[draw=none,fill=color0] (axis cs:-0.767656558881948,0) rectangle (axis cs:-0.575742419161461,223);
\draw[draw=none,fill=color0] (axis cs:-0.575742419161461,0) rectangle (axis cs:-0.383828279440974,234);
\draw[draw=none,fill=color0] (axis cs:-0.383828279440974,0) rectangle (axis cs:-0.191914139720487,61);
\draw[draw=none,fill=color0] (axis cs:-0.191914139720487,0) rectangle (axis cs:0,9);
\draw[draw=none,fill=color0] (axis cs:0,0) rectangle (axis cs:0.191914139720487,10);
\draw[draw=none,fill=color0] (axis cs:0.191914139720487,0) rectangle (axis cs:0.383828279440974,63);
\draw[draw=none,fill=color0] (axis cs:0.383828279440974,0) rectangle (axis cs:0.575742419161461,182);
\draw[draw=none,fill=color0] (axis cs:0.575742419161461,0) rectangle (axis cs:0.767656558881948,191);
\draw[draw=none,fill=color0] (axis cs:0.767656558881948,0) rectangle (axis cs:0.959570698602435,142);
\draw[draw=none,fill=color0] (axis cs:0.959570698602435,0) rectangle (axis cs:1.15148483832292,112);
\draw[draw=none,fill=color0] (axis cs:1.15148483832292,0) rectangle (axis cs:1.34339897804341,114);
\draw[draw=none,fill=color0] (axis cs:1.34339897804341,0) rectangle (axis cs:1.5353131177639,51);
\draw[draw=none,fill=color0] (axis cs:1.5353131177639,0) rectangle (axis cs:1.72722725748438,39);
\draw[draw=none,fill=color0] (axis cs:1.72722725748438,0) rectangle (axis cs:1.91914139720487,35);
\draw[draw=none,fill=color0] (axis cs:1.91914139720487,0) rectangle (axis cs:2.11105553692536,21);
\draw[draw=none,fill=color0] (axis cs:2.11105553692536,0) rectangle (axis cs:2.30296967664584,14);
\draw[draw=none,fill=color0] (axis cs:2.30296967664584,0) rectangle (axis cs:2.49488381636633,4);
\draw[draw=none,fill=color0] (axis cs:2.49488381636633,0) rectangle (axis cs:2.68679795608682,3);
\draw[draw=none,fill=color0] (axis cs:2.68679795608682,0) rectangle (axis cs:2.87871209580731,2);
\end{axis}

\end{tikzpicture}
	\caption{Histograms showing the final distribution of the $\lambda$s from Plot~\ref{fig:init_lambda}. As the optimization process converges, we observe significant movement away from the center towards the sides.}
	\label{fig:distribution_lambda}
\end{figure}
Standard SGD worked better with initialization to ones. The Adam-optimizer, however, works with the standard normal initialization. 
Results are shown in Table~\ref{tab:init_alexnet_ADAM}. We observe a slightly improved performance in comparison to what we saw in Table \ref{tab:init_alexnet}. Previously initialization by running the power method worked slightly better. Now the maximum values are higher for the normally or uniformly initialized CP-forms. Since the standard deviations indicate that the differences are not significant, we conclude that running the power method is not required when using Adam. The progression of $\lambda$ distributions during training is shown in Figures \ref{fig:init_lambda} and \ref{fig:distribution_lambda}. Encouragingly Adam pushes the scales towards the sides away from zero, as we would expect in a working system.

\subsection{Network Compression}\label{sec:compression}
Re-parameterizing the network in a CP-form, allows easy compression by truncating the rank sums. We compress our networks by removing the smallest rank scale lambdas and their corresponding vectors from the decomposition.
The following experiments were conducted on the AlexNet and LeNet-like architectures (as discussed in the previous section) with various compression rates.

\subsubsection{AlexNet-CIFAR10}\label{sec:alex_cifar_compression}
\begin{table}[t]
	\centering
	\caption{Sum truncation for all layers of the AlexNet-like architecture on CIFAR-10. Various compression rates and the resulting accuracies are tabulated.
	The compressed networks have been fine-tuned using SGD with variety of different learning rates, which are specified in brackets.}
	\label{tab:AlexNet-CIFAR10-compression}
	\begin{tabular}{ c c c c c } \toprule
		 			&				& \multicolumn{2}{c}{accuracy [\%]}  &			  \\ \cmidrule{3-4}
		 compression&  learning rate&  max  & $\mu \pm \sigma$ & \# weights			  \\ \midrule
		0\%   		& 0.001   		& 89.05& $88.32\pm 0.21$ & $\; 9.25\times 10^6$ \\  
		25\%		& 0.0001     	& 89.12& $89.10\pm 0.02$ & $\; 6.93\times 10^6$ \\ 
		50\% 		& 0.001  		& 87.93& $87.79\pm 0.05$ & $\; 4.62\times 10^6$ \\ 
		75\% 		& 0.01  		& 81.61& $81.25\pm 0.28$ & $\; 2.31\times 10^6$ \\ \bottomrule
	\end{tabular}
\end{table}

\begin{table}[t]
	\centering
	\caption{Compression performance comparison between Tai compression and CP-compression. Here, we only consider compression of the convolutional layers and give the number of weights for these.
	All the convolutional layers in AlexNet except the first one are compressed. Compression is performed by fine tuning over 20 epochs.
	This is an important difference to the experiments run for table~\ref{tab:AlexNet-CIFAR10-compression}.}
	\label{tab:Tai-compare-compression}
	\begin{tabular}{c c c c}\toprule
		method   							 & compression 		&  accuracy	[\%]	    & 	\# weights      		 \\ \midrule
		low-rank \cite{Tai2016Convolutional} &		0\%			&  85.11		    & 	$2.25\times 10^6$	 \\
		canonical (ours) 					 &		0\%			&  85.96			&	$3.21\times 10^6$    \\
		low-rank \cite{Tai2016Convolutional} &		89.78\%		&  $83.06\pm 1.02$  &   $0.23\times 10^6$    \\
		canonical (ours)     				 & 		90.91\%   	&  $81.82\pm 0.52$  &	$0.20\times 10^6$    \\ \bottomrule
	\end{tabular}
\end{table}

Before compression, the full-rank networks are trained and stored.
We measure compression with respect to the best fit CP-normalized network. 0\% means that we are working with every outer product. 
We compress our networks by sorting the CP-summands for each rank according to their $\lambda_r$ weights. After sorting, we discard, for example, the 25\% least important CP-summands. All summands have the same amount of parameters. Therefore, in this case, 75\% of parameters are retained.

The compressed networks are fine-tuned for 20 epochs with SGD to compensate for the truncation. We use SGD for all fine-tuning as RMSprop was unstable in some cases. During tuning, we observed over-fitting problems depending on the learning rates.
In response, we chose the learning rates based on the validation accuracy after 20 epochs. 
For twenty-five percent compression, we work with a learning rate of 0.0001, which is much smaller than the optimization step size of the initial training. 
In the 50\% compression case, a learning rate of 0.001 is used. 
For 75\% compression, we chose an even higher learning rate of 0.01 for faster convergence. As the distance to the original network weights increases, learning rates become useful. We find this relationship intuitive since the distance we must travel in weight space to compensate for the missing parameters increases.

Table~\ref{tab:AlexNet-CIFAR10-compression} contains the compression results for our AlexNet-like structure. 
The number of parameters at 25\% compression and the corresponding accuracy is particularly significant. Here, the number of parameters approximately equals those of the weight-normalized network. The performance improves in comparison to the full rank or perfect fit parametrization, Table~\ref{tab:cifar10_alexnet}. We conclude that a near-perfect fit is not required in this case.

With half of the parameters, we observe a $\approx 1\%$ accuracy drop compared to the full rank CP-normalized network. Finally, we cut the CP-sum short after the first quarter, effectively removing 75\% of all parameters. At the same time, this drastic parameter cut results in only a $\approx 7\%$ accuracy drop.

Note, the 25\% and 50\% compressed networks outperform the weight normalized and un-normalized networks shown in Table~\ref{tab:cifar10_alexnet} in terms of mean accuracy, the version with only 50\% of the CP-summands does so with significantly fewer parameters.  

\subsubsection{Comparison to Tai et al. \cite{Tai2016Convolutional}}
Table~\ref{tab:Tai-compare-compression} compares the compression performance of the CP-form and the low rank-formulation proposed in \cite{Tai2016Convolutional}. We work with the AlexNet-like network from section~\ref{sec:alex_cifar_compression} on CIFAR10.
To challenge both methods, we aim for compression rates of approximately $90\%$. 
During the fine-tuning \cite{Tai2016Convolutional} employs batch normalization, while our CP-form does not.
We observe competitive performance for our approach, with slightly fewer parameters than our re-implementation of \cite{Tai2016Convolutional}.
The method proposed in this paper allows compression of convolutional and fully connected layers. Since \cite{Tai2016Convolutional} do not consider dense layers, we limit ourselves to convolutional layers here for a fair comparison.
As \cite{Tai2016Convolutional} chose to work with Lua, we re-implemented their approach in PyTorch. Our source code is made available.

\subsubsection{Compression Performance on SVHN}
We now move to the compression of the network resulting from our SVHN experiments.
\begin{table}[t]
	\centering
	\caption{Compression results on AlexNet like architecture on SVHN with various compression rates using a SGD optimizer with variety of learning rates specified in brackets.}\label{tab:alexnet-svhn-compression}
	\begin{tabular}{ c c c c c } \toprule
					 &				 & \multicolumn{2}{c}{accuracy [\%]} &   \\ \cmidrule{3-4}
		compression  & learning rate & max   & $\mu \pm \sigma$ &	\# weights \\ \midrule
		 0\% 		 & 0.0001 		 & 95.62 & $94.82\pm 0.30$ & $\; 9.25\times 10^6$ \\  
		25\% 		 & 0.0001 	     & 95.30 & $95.27\pm 0.01$ & $\; 6.93\times 10^6$ \\ 
		50\% 		 & 0.001  		 & 95.42 & $95.36\pm 0.03$ & $\; 4.62\times 10^6$ \\ 
		75\% 		 & 0.01   		 & 94.19 & $93.75\pm 0.14$ & $\; 2.31\times 10^6$ \\ \bottomrule
	\end{tabular}
\end{table}
Accuracies, compression-rates as well as parameter counts appear in Table~\ref{tab:alexnet-svhn-compression}.
We use the same compression rates as in the case of CIFAR10 and again fine-tune for 20 epochs. Once more, we find increasing learning rates more helpful when networks are compressed more aggressively. 

On SVHN, we find that removing the last quarter and half of the CP-sum improves our mean results. Going further and removing three-quarters of the network parameters has a detrimental effect on the network accuracy. We deduce that the network initially had more parameters than required. We can classify the SVHN-digits with fewer weights. In this case, we settle with the upper half of the original summands.

The best performing 50\% CP-normalized network improves upon the weight normalized and not normalized networks we saw in Table~\ref{tab:SVHN_alexnet} in terms of both mean accuracy and parameters.

\subsubsection{LeNet-MNIST}
\begin{table}[t]
	\centering
	\caption{Compression results on LeNet like architecture with various compression rates using a RMSprop optimizer with a learning rate of 0.001.}\label{tab:LeNet-Mnist-compression}
	\begin{tabular}{ c c c c } \toprule
		        & \multicolumn{2}{c}{accuracy [\%]}  &   \\ \cmidrule{2-3}
	compression	& max & $\mu \pm \sigma$ & \# weights \\ \midrule
	0\%                &  99.35    & $99.21\pm 0.05$ & $\; 1.22\times 10^6$ \\ 
	10\%               &  99.23    & $99.18\pm 0.01$ & $\; 1.10\times 10^6$ \\ 
	25\%               &  98.83    & $98.80\pm 0.02$ & $\; 0.91\times 10^6$ \\ 
	50\%               &  97.56    & $97.55\pm 0.02$ & $\; 0.61\times 10^6$ \\ \bottomrule
	\end{tabular}
\end{table}
	
Finally, we compress the LeNet-like architecture we trained on MNIST with RMSprop and a learning rate of 0.001.

We remove the lower 10\%, 25\%, and 50\% of the summands in the CP-form and collect the resulting classification accuracies in Table~\ref{tab:LeNet-Mnist-compression}. At a 10\% compression rate, the resulting mean accuracy drop we observe is fairly small 0.01\% . At 25\% compression, a comparatively small drop of $\approx 0.5\%$. In this case, we can not afford to discard half of the parameters since we find the performance loss significant.

\section{Conclusion}\label{sec:conclusions}
In this paper, we proposed to express network weights as a weighted sum of normalized outer vector products. 
Computing a canonical/parallel factor decomposition allows a direct network initialization using well-known, established approaches.
We evaluated the newly formulated method experimentally and found it to rival weight normalization in terms of network convergence and stability.
In contrast to weight normalization, the canonical approach simplifies network compression after training. We tested initializing the CP-form directly. Our evidence suggests that standard initialization techniques can replace the power method for network initialization.

\subsection{Future Work}
In hindsight, we would have additionally included lower ranks during our initial training. Given the encouraging compression results, this is certainly an idea we recommend for future work. An adaptive compression based on the size of the rank weights $\lambda_r$ should also be investigated.

\bibliographystyle{plain}
\bibliography{bib}

\end{document}


\title{Canonical convolutional neural networks: Supplementary Material}

\maketitle

\section*{S1  Rank Estimation}\label{sec:intro}
\begin{figure}[h!]
	\centering
		\includegraphics[width=\linewidth, height=8.75cm]{./figures/fit_plot}
	\caption{Figure illustrating the improvement in reconstructed tensor fit with respect to rank of the tensor with shape 64x192x3x3. 
    Fit is calculated using the orignal tensor $\mathcal{W}$ and $\bar{\mathcal{W}}$} \label{fig:rank_plot}
\end{figure}
For the application of CP normalization we need a prior knowledge of the full rank of every tensor.
Figure~\ref{fig:rank_plot} shows the improvement in reconstructed tensor from factor matrices with 
increase in rank for a second layer of AlexNet like architecture with shape of 64x192x3x3. 
We observe linear growth in fit percentage with increase in tensor.
All the decompositions are carried out using the parafac decompositon from tensorly.
Reconstructed tensor is computed using the cp\_to\_tensor method from tensorly.